\documentclass[11pt,letterpaper]{article}
\usepackage[pass]{geometry}
\usepackage{emnlp2017}
\usepackage{times}
\usepackage{latexsym}
\usepackage{url}
\usepackage{graphicx}
\usepackage{subcaption}
\usepackage{amssymb}
\usepackage{amsfonts}
\usepackage{amsmath}
\usepackage{graphicx}
\usepackage{xspace}
\usepackage{xcolor}
\usepackage{pifont}
\usepackage{enumitem}
\usepackage{url}
\usepackage{latexsym}
\usepackage{array}
\usepackage{multirow}
\usepackage{arydshln}
\usepackage{array}
\graphicspath{ {./}}

\emnlpfinalcopy

\newcolumntype{L}[1]{>{\raggedright\let\newline\\\arraybackslash\hspace{0pt}}m{#1}}
\newcolumntype{C}[1]{>{\centering\let\newline\\\arraybackslash\hspace{0pt}}m{#1}}
\newcolumntype{R}[1]{>{\raggedleft\let\newline\\\arraybackslash\hspace{0pt}}m{#1}}

\newcommand{\pca}{\textsc{pca}\xspace}

\newcommand{\tbase}{\textit{t}\textsc{base}\xspace}
\newcommand{\ubase}{\textit{u}\textsc{base}\xspace}
\newcommand{\rnn}{\textsc{rnn}\xspace}

\newcommand{\uernn}{\textit{ue}\textsc{rnn}\xspace}
\newcommand{\ternn}{\textit{te}\textsc{rnn}\xspace}
\newcommand{\ubrnn}{\textit{ub}\textsc{rnn}\xspace}
\newcommand{\tbrnn}{\textit{tb}\textsc{rnn}\xspace}

\newcommand{\arnn}{\textit{a}-\textsc{rnn}\xspace}

\newcommand{\cnn}{\textsc{cnn}\xspace}

\newcommand{\mlp}{\textsc{mlp}\xspace}
\newcommand{\lr}{\textsc{lr}\xspace}
\newcommand{\WordVec}{\textsc{word2vec}\xspace}

\newcommand{\GazzettaTrain}{\textsc{g-train}\xspace}

\newcommand{\GazzettaDev}{\textsc{g-dev}\xspace}
\newcommand{\GazzettaTest}{\textsc{g-test}\xspace}

\newcommand{\R}{\mathbb{R}}
\newcommand{\Detox}{\textsc{detox}\xspace}
\newcommand{\gru}{\textsc{gru}\xspace}

\newcommand{\auc}{\textsc{auc}\xspace}

\newcommand{\svm}{\textsc{svm}\xspace}
\newcommand{\id}{\textsc{id}\xspace}
\title{Improved Abusive Comment Moderation with User Embeddings}

\author{
	John Pavlopoulos\\ \textbf{Prodromos Malakasiotis} \\ \textbf{Juli Bakagianni} \\ 
	Straintek, Athens, Greece \\{\tt \{ip, mm, jb\}@straintek.com} \\\And
	Ion Androutsopoulos \\
	Department of Informatics \\
	Athens University of Economics \\and Business, Greece \\
	{\tt ion@aueb.gr} \\
}
  
\date{}
\hypersetup{draft}

\begin{document}
\maketitle
\begin{abstract}
Experimenting with a dataset of approximately 1.6M user comments from a Greek news sports portal, we explore how a state of the art \rnn-based moderation method can be improved by adding user embeddings, user type embeddings, user biases, or user type biases. We observe improvements in all cases, with user embeddings leading to the biggest performance gains.
\end{abstract}


\section{Introduction}

News portals often allow their readers to comment on articles, in order to get feedback, engage their readers, and build customer loyalty. User comments, however, can also be abusive (e.g., bullying, profanity, hate speech), damaging the reputation of news portals, making them liable to fines (e.g., when hosting comments encouraging illegal actions),  and putting off readers. Large news portals often employ moderators, who are frequently overwhelmed by the volume and abusiveness of comments.\footnote{See, for example, \url{https://goo.gl/WTQyio}.} Readers are disappointed when non-abusive comments do not appear quickly online because of moderation delays. Smaller news portals may be unable to employ moderators, and some are forced to shut down their comments.\footnote{See \url{https://goo.gl/2eKdeE}.} 

In previous work \cite{Pavlopoulos2017}, we introduced a new dataset of approx.\ 1.6M manually moderated user comments from a Greek sports news portal, called Gazzetta, which we made publicly available.\footnote{The portal is \url{http://www.gazzetta.gr/}. Instructions to download the dataset will become available at \url{http://nlp.cs.aueb.gr/software.html}.} Experimenting on that dataset and the datasets of Wulczyn et al.\ \shortcite{Wulczyn2017}, which contain moderated English Wikipedia comments, we  showed that a method based on a Recurrent Neural Network (\rnn) outperforms \Detox \cite{Wulczyn2017}, the previous state of the art in automatic user content moderation.\footnote{Two of the co-authors of Wulczyn et al.\ \shortcite{Wulczyn2017} are with Jigsaw, who recently announced Perspective, a system to detect ‘toxic’ comments. Perspective is not the same as  \Detox (personal communication), but we were unable to obtain scientific articles describing it.} Our previous work, however, considered only the texts of the comments, ignoring user-specific information (e.g., number of previously accepted or rejected comments of each user). Here we add \emph{user embeddings} or \emph{user type embeddings} to our \rnn-based method, i.e., dense vectors that represent individual users or user types, similarly to word embeddings that represent words \cite{Mikolov2013c,Pennington2014}. Experiments on Gazzetta comments show that both user embeddings and user type embeddings improve the performance of our \rnn-based method, with user embeddings helping more. User-specific or user-type-specific \emph{scalar biases} also help to a lesser extent. 


\section{Dataset}

\begin{table*}
\centering
{\small
\begin{tabular}{|c|c|c|c|c|c|c|c|}
\hline
\multirow{2}{*}{Dataset/Split} & \multicolumn{2}{|c|}{Gold Label} & \multicolumn{4}{|c|}{Comments Per User Type} & \multirow{2}{*}{Total} \\\cline{2-7}
                & Accepted       & Rejected       & Green          & Yellow         & Red          & Unknown       & \\\hline
\GazzettaTrain  & 960,378 (66\%) & 489,222 (34\%) & 724,247 (50\%) & 585,622 (40\%) & 43,702 (3\%) & 96,029 (7\%)  & 1.45M \\
\GazzettaDev    & 20,236 (68\%)  & 9,464 (32\%)   & 14,378 (48\%)  & 10,964 (37\%)  & 546 (2\%)    & 3,812 (13\%)  & 29,700 \\
\GazzettaTest   & 20,064 (68\%)  & 9,636 (32\%)   & 14,559 (49\%)  & 10,681 (36\%)  & 621 (2\%)    & 3,839 (13\%)  & 29,700 \\\hline
\end{tabular}
}
\caption{Comment statistics of the dataset used.}
\vspace{-4mm}
\label{tab:data_stats}
\end{table*}

\begin{table}
\centering
{\small
\begin{tabular}{|c|c|c|c|c|c|c|c|}
\hline
\multirow{2}{*}{Dataset/Split} & \multicolumn{4}{|c|}{Individual Users Per User Type} & \multirow{2}{*}{Total} \\\cline{2-5}
	           & Green  & Yellow & Red & Unknown                & \\\hline
\GazzettaTrain & 4,451  & 3,472  & 251 & 21,865 $\rightarrow$ 1 & 8,175 \\
\GazzettaDev   & 1,631  & 1,218  & 64  & 1,281 $\rightarrow$ 1  & 2,914 \\
\GazzettaTest  & 1,654  & 1,203  & 67  & 1,254 $\rightarrow$ 1  & 2,925 \\\hline
\end{tabular}
}
\caption{User statistics of the dataset used.}
\vspace{-5mm}
\label{tab:user_stats}
\end{table}

We first discuss the dataset we used, to help acquaint the reader with the problem. The dataset contains Greek comments from Gazzetta \cite{Pavlopoulos2017}. There are approximately 1.45M training comments (covering Jan.\ 1, 2015 to Oct.\ 6, 2016); we call  them \GazzettaTrain (Table~\ref{tab:data_stats}). An additional set of 60,900 comments (Oct.\ 7 to Nov.\ 11, 2016) was split to development set (\GazzettaDev, 29,700 comments) and test set  (\GazzettaTest, 29,700).\footnote{The remaining 1,500 comments are not used here. Smaller subsets of \GazzettaTrain and \GazzettaTest are also available \cite{Pavlopoulos2017}, but are not used in this paper. The Wikipedia comment datasets of Wulczyn et al.\ \shortcite{Wulczyn2017} cannot be used here, because they do not provide user \id{s}.} Each comment has a gold label (`accept', `reject'). The user \id of the author of each comment is also available, but user \id{s} were not used in our previous work. 

When experimenting with \emph{user type} embeddings or biases, we group the users into the following types. $T(u)$ is the number of training comments posted by user (\id) $u$. $R(u)$ is the ratio of training comments posted by $u$ that were rejected.

\smallskip

\noindent \textbf{Red:} Users with $T(u) > 10$ and $R(u) \geq 0.66$.

\noindent \textbf{Yellow:} $T(u) > 10$ and $0.33 < R(u) < 0.66$.

\noindent \textbf{Green:} $T(u) > 10$ and $R(u) \leq 0.33$.

\noindent \textbf{Unknown:} Users with $T(u) \leq 10$. 

\smallskip
\noindent  Table~\ref{tab:user_stats} shows the number of users per type. 


\section{Methods} \label{sec:methods}

\noindent\textbf{\rnn:} This is the \rnn-based method of our previous work \cite{Pavlopoulos2017}. It is a chain of \gru cells \cite{Cho2014} that transforms the tokens $w_{1} \dots, w_{k}$ of each comment to the hidden states $h_{1} \dots, h_{k}$ ($h_i \in \R^m$). Once $h_k$ has been computed, a logistic regression (\lr) layer estimates the probability that comment $c$ should be rejected: 
\begin{equation}
P_{\rnn}(\textit{reject}|c) = \sigma(W_{p} h_{k} + b) 
\end{equation}
$\sigma$ is the sigmoid function, $W_p \in \R^{1 \times m}$, $b \in \R$.\footnote{In our previous work \cite{Pavlopoulos2017}, we also considered a variant of \rnn, called \arnn, with an attention mechanism. We do not consider \arnn here to save space.} 

\smallskip
\noindent \textbf{\uernn:} This is the \rnn-based method with \emph{user embeddings} added. Each user $u$ of the training set with $T(u) > 10$ is mapped to a user-specific embedding $v_u \in \R^{d}$. Users with $T(u) \leq 10$ are mapped to a single `unknown' user embedding. The \lr layer is modified as follows; $v_u$ is the embedding of the author of $c$; and $W_v \in \R^{1 \times d}$.
\begin{equation}
P_{\uernn}(\textit{reject}|c) = \sigma(W_{p} h_{k} + W_{v} v_u + b) 
\label{eq:uernn}
\end{equation}

\noindent \textbf{\ternn:} This is the \rnn-based method with \emph{user type embeddings} added. Each user type $t$  is mapped to a user type embedding $v_t \in \R^{d}$. The \lr layer is modified as follows, where $v_t$ is the embedding of the type of the author of $c$. 
\begin{equation}
P_{\ternn}(\textit{reject}|c) = \sigma(W_{p} h_{k} + W_{v} v_t + b) 
\end{equation}

\noindent \textbf{\ubrnn:} This is the \rnn-based method with \emph{user biases} added. Each user $u$ of the training set with $T(u) > 10$ is mapped to a user-specific bias $b_u \in \R$. Users with $T(u) \leq 10$ are mapped to a single `unknown' user bias. The \lr layer is modified as follows, where $b_u$ is the bias of the author of $c$. 
\begin{equation}
P_{\ubrnn}(\textit{reject}|c) = \sigma(W_{p} h_{k} + b_{u}) 
\end{equation}
We expected \ubrnn to learn higher (or lower) $b_u$ biases for users whose posts were frequently rejected (accepted) in the training data, biasing the system towards rejecting (accepting) their posts. 

\smallskip
\noindent \textbf{\tbrnn:} This is the \rnn-based method with \emph{user type biases}. Each user type $t$  is mapped to a user type bias $b_t \in \R$. The \lr layer is modified as follows; $b_t$ is the bias of the type of the author.
\begin{equation}
P_{\tbrnn}(\textit{reject}|c) = \sigma(W_{p} h_{k} + b_{t}) 
\label{eq:tbrnn}
\end{equation}
We expected \tbrnn to learn a higher $b_t$ for the red user type (frequently rejected), and a lower $b_t$ for the green user type (frequently accepted), with the biases of the other two types in between.

In all methods above, we use 300-dimensional word embeddings, user and user type embeddings with $d = 300$ dimensions, and $m=128$ hidden units in the \gru cells, as in our previous experiments \cite{Pavlopoulos2017}, where we tuned all hyper-parameters on 2\% held-out training comments. Early stopping evaluates on the same held-out subset.
User and user type embeddings are randomly initialized and updated by backpropagation. Word embeddings are initialized to the \WordVec embeddings of our previous work \cite{Pavlopoulos2017}, which were pretrained on 5.2M Gazzetta comments. Out of vocabulary words, meaning words not encountered or encountered only once in the training set and/or words with no initial embeddings, are mapped (during both training and testing) to a single randomly initialized word embedding, updated by backpropagation. We use Glorot initialization \cite{Glorot2010} for other parameters, cross-entropy loss, and Adam \cite{Kingma2015}.\footnote{We used Keras (\url{http://keras.io/}) with the TensorFlow back-end (\url{http://www.tensorflow.org/}).}

\smallskip
\noindent \textbf{\ubase:} For a comment $c$ authored by user $u$, this baseline returns the rejection rate $R(u)$ of the author's training comments, if there are $T(u) > 10$ training comments of $u$, and 0.5 otherwise. 
\[
P_{\ubase}(\textit{reject}|c) =
\left\{
\begin{array}{l}
R(u), \textrm{ if } T(u) > 10\\
0.5, \textrm{ if } T(u) \leq 10
\end{array}
\right.
\]

\noindent \textbf{\tbase:} This baseline returns the following probabilities, considering the user type $t$ of the author.
\[
P_{\tbase}(\textit{reject}|c) =
\left\{
\begin{array}{l}
1, \textrm{if $t$ is Red}\\
0.5, \textrm{if $t$ is Yellow}\\
0.5, \textrm{if $t$ is Unknown}\\
0, \textrm{if $t$ is Green}
\end{array}
\right.
\]


\section{Results and Discussion} \label{sec:results}

\begin{table}
\centering
{\small
\begin{tabular}{|c|c|c|}
\hline
System  & \GazzettaDev                           & \GazzettaTest \\\hline
\uernn & $\mathbf{80.68 \left(\pm 0.11\right)}$ & $\mathbf{80.71 \left(\pm 0.13\right)}$\\\hline
\ubrnn   & $80.54 \left(\pm 0.09\right)$          & $80.53 \left(\pm 0.08\right)$ \\\hline
\ternn & $80.37 \left(\pm 0.05\right)$ & $80.41 \left(\pm 0.09\right)$ \\\hline
\tbrnn & $80.33 \left(\pm 0.12\right)$ & $80.32 \left(\pm 0.05\right)$\\\hline
\rnn    & $79.40 \left(\pm 0.08\right)$          & $79.24 \left(\pm 0.05\right)$ \\\hline
\ubase & $67.61$ & $68.57$ \\\hline
\tbase & $63.16$ & $63.82$ \\\hline
\end{tabular}
}
\caption{\auc scores. Standard error in brackets.}
\vspace{-5mm}
\label{tab:eval_scores}
\end{table}

Table~\ref{tab:eval_scores} shows the \auc scores (area under \textsc{roc} curve) of the methods considered. Using \auc allows us to compare directly to the results of our previous work \cite{Pavlopoulos2017} and the work of Wulczyn et al.\ \shortcite{Wulczyn2017}. Also, \auc considers performance at multiple classification thresholds $t$ (rejecting comment $c$ when $P(\textit{reject}|c) \geq t$,  for different $t$ values), which gives a more complete picture compared to reporting precision, recall, or F-scores for a particular $t$ only. Accuracy is not an appropriate measure here, because of class imbalance (Table~\ref{tab:data_stats}).  For methods that involve random initializations (all but the baselines), the results are averaged over three repetitions; we also report the standard error across the repetitions.  

\begin{figure}
	\centering
	\includegraphics[width=\columnwidth]{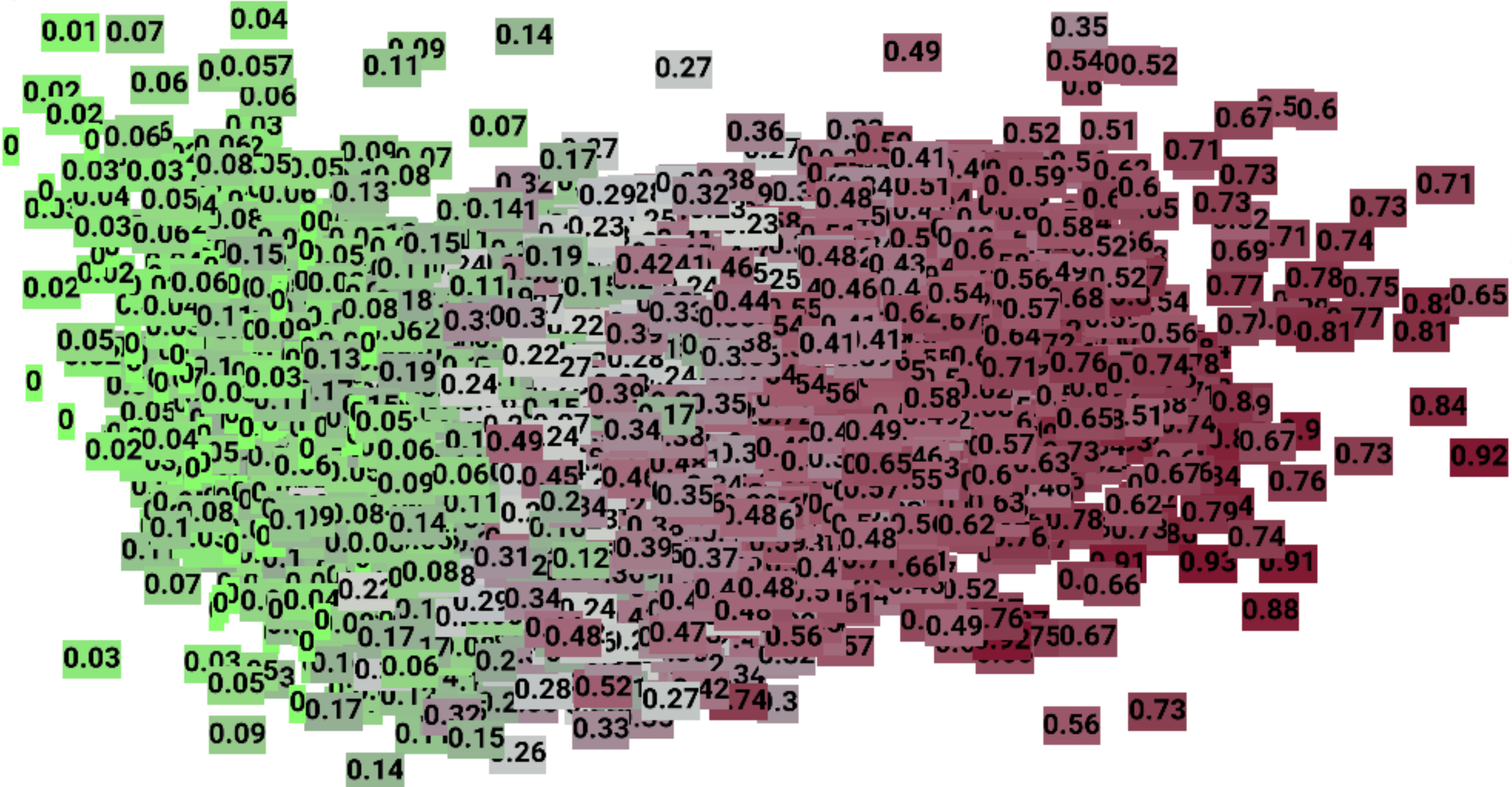}
	\caption{User embeddings learned by \uernn (2 principal components). Color represents the rejection rate $R(u)$ of the user's training comments. }
	\label{fig:user_zones}
\end{figure}

User-specific information always improves our original \rnn-based method  (Table~\ref{tab:eval_scores}), but the best results are obtained by adding user embeddings (\uernn). Figure~\ref{fig:user_zones} visualizes the user embeddings learned by \uernn. The two dimensions of Fig.~\ref{fig:user_zones} correspond to the two principal components of the user embeddings, obtained via \pca.The colors and numeric labels reflect the rejection rates $R(u)$ of the corresponding users. Moving from left to right in Fig.~\ref{fig:user_zones}, the rejection rate increases, indicating that the user embeddings of \uernn capture mostly the rejection rate $R(u)$. This rate (a single scalar value per user) can also be captured by the simpler user-specific biases of \ubrnn, which explains why \ubrnn also performs well (second best results in Table~\ref{tab:eval_scores}). Nevertheless, \uernn performs better than \ubrnn, suggesting that user embeddings capture more information than just a user-specific rejection rate bias.\footnote{We obtained no clear clusterings with tSNE \cite{Maaten2008}.
}

Three of the user types (Red, Yellow, Green) in effect also measure $R(u)$, but in discretized form (three bins), which also explains why user type embeddings (\ternn) also perform well (third best method). The performance of  \tbrnn is close to that of \ternn, suggesting again that most of the information captured by user type embeddings can also be captured by simpler scalar user-type-specific biases. The user type biases $b_t$ learned by \tbrnn are shown in Table~\ref{tab:bt_values}. The bias of the Red type is the largest, the bias of the Green type is the smallest, and the biases of the Unknown and Yellow types are in between, as expected (Section~\ref{sec:methods}). The same observations hold for the average user-specific biases $b_u$ learned by \ubrnn (Table~\ref{tab:bt_values}).

Overall, Table~\ref{tab:eval_scores} indicates that user-specific information (\uernn, \ubrnn) is better than user-type information (\ternn, \tbrnn), and that embeddings (\uernn, \ternn) are better than the scalar biases (\ubrnn, \tbrnn), though the differences are small. All the \rnn-based methods outperform the two baselines (\ubase, \tbase), which do not consider the texts of the comments.
 
Let us provide a couple of examples, to illustrate the role of user-specific information. We encountered a comment saying just ``Ooooh, down to Pireaus\dots'' (translated from Greek), which the moderator had rejected, because it is the beginning of an abusive slogan. The rejection probability of \rnn was only 0.34, presumably because there are no clearly abusive expressions in the comment, but the rejection probability of \uernn was 0.72, because the author had a very high rejection rate. On the other hand, another comment said ``Indeed, I know nothing about the filth of Greek soccer.'' (translated, apparently not a sarcastic comment). The original \rnn method marginally rejected the comment (rejection probability 0.57), presumably because of the `filth' (comments talking about the filth of some sport or  championship are often rejected), but \uernn gave it a very low rejection probability (0.15), because the author of the comment had a very low rejection rate. 


\section{Related work} \label{sec:relatedWork}

In previous work \cite{Pavlopoulos2017}, we showed that our \rnn-based method outperforms \Detox \cite{Wulczyn2017}, the previous state of the art in user content moderation. \Detox uses character or word $n$-gram features, no user-specific information, and 
an \lr or \mlp classifier. Other related work on abusive content moderation was reviewed extensively in our previous work \cite{Pavlopoulos2017}. Here we focus on previous work that considered user-specific 
features and user embeddings.

\begin{table}
\centering
{\small
\begin{tabular}{|c|c|c|}
\hline
User Type    & $b_t$  of \tbrnn & average $b_u$ of \ubrnn \\\hline
Green        & $-0.471 \left(\pm 0.007\right) $ & $-0.180 \left(\pm 0.024\right) $ \\\hline
Yellow       & $0.198 \left(\pm 0.015\right) $ & $0.058 \left(\pm 0.022\right) $  \\\hline
Unknown      & $0.256 \left(\pm 0.021\right) $ & $0.312 \left(\pm 0.011\right) $ \\\hline
Red          & $1.151 \left(\pm 0.013\right) $ & $0.387 \left(\pm 0.023\right) $ \\\hline
\end{tabular}
}
\caption{Biases learned and standard error.}
\vspace{-5mm}
\label{tab:bt_values}
\end{table}

Dadvar et al.\ \shortcite{Dadvar2013} detect cyberbullying in YouTube comments, using 
an \svm and features examining the content of each comment (e.g., second person pronouns followed by profane words, common bullying words), but also the profile and history of the author of the comment (e.g., age, frequency of profane words in past posts). 
Waseem et al.\ \shortcite{Waseem2016} detect hate speech tweets. Their best method is an \lr classifier, with character $n$-grams and a feature indicating the gender of the author; adding the location of the author did not help. 

Cheng et al.\ \shortcite{Cheng2015} predict which users will be banned from on-line communities. Their best
system uses a Random Forest or \lr classifier, with features examining the average readability and sentiment of each user's past posts, the past activity of each user (e.g., number of posts daily, proportion of posts that are replies), and the reactions of the community to the past actions of each user (e.g., up-votes, number of posts rejected).
Lee et al.\ \shortcite{Lee2014} and Napoles et al.\ \shortcite{Napoles2017b} include similar user-specific features in classifiers intended to detect high quality on-line discussions.

Amir et al.\ \shortcite{Amir2016} detect sarcasm in tweets. Their best system uses a word-based Convolutional Neural Network (\cnn).
The feature vector produced by the \cnn (representing the content of the tweet) is concatenated with the user embedding of the author, and passed on to an \mlp 
that classifies the tweet as sarcastic or not. 
This method outperforms a previous state of the art sarcasm detection method \cite{Bamman2015} that relies on an \lr classifier with 
hand-crafted content and user-specific features. 
We use an \rnn instead of a \cnn, and we feed the comment and user embeddings to a simpler \lr layer (Eq.~\ref{eq:uernn}), instead of an \mlp. 
Amir et al.\ discard unknown users, unlike our experiments, and consider only sarcasm, whereas moderation also involves 
profanity, hate speech, bullying, threats etc. 

User embeddings have also been used in: conversational agents \cite{Li2016}; sentiment analysis \cite{Chen2016}; retweet prediction \cite{Zhang2016}; predicting which topics a user is likely to tweet about, the accounts a user may want to follow, and the age, gender, political affiliation of Twitter users \cite{Benton2016}. 

Our previous work \cite{Pavlopoulos2017} also discussed how machine learning can be used in \emph{semi-automatic} moderation, by letting moderators focus on `difficult' comments and automatically handling comments that are easier to accept or reject. In more recent work \cite{Pavlopoulos2017emnlp} we also explored how an attention mechanism can be used to highlight possibly abusive words or phrases when showing `difficult' comments to moderators.


\section{Conclusions} \label{sec:conc}

Experimenting with a dataset of approx.\ 1.6M user comments from a Greek sports news portal, we explored how a state of the art \rnn-based moderation method can be improved by adding user embeddings, user type embeddings, user biases, or user type biases. We observed improvements in all cases, but user embeddings were the best.

We plan to compare \uernn to \cnn-based methods that employ user embeddings 
\cite{Amir2016}, after replacing the \lr layer of \uernn by an \mlp to allow non-linear combinations of comment and user embeddings.


\section*{Acknowledgments}

This work was funded by Google's Digital News Initiative (project \textsc{ml2p}, contract 362826).\footnote{See \url{https://digitalnewsinitiative.com/}.} We are grateful to Gazzetta for the data they provided. We also thank Gazzetta's moderators for their feedback, insights, and advice.

\bibliography{references}

\begin{thebibliography}{20}
\expandafter\ifx\csname natexlab\endcsname\relax\def\natexlab#1{#1}\fi

\bibitem[{Amir et~al.(2016)Amir, Wallace, Lyu, Carvalho, and
  M.~J.~Silva}]{Amir2016}
S.~Amir, B.~C. Wallace, H.~Lyu, P.~Carvalho, and Mario~J. M.~J.~Silva. 2016.
\newblock Modelling context with user embeddings for sarcasm detection in
  social media.
\newblock In \emph{Proceedings of CoNLL}, pages 167--177, Berlin, Germany.

\bibitem[{Bamman and Smith(2015)}]{Bamman2015}
D.~Bamman and N.A. Smith. 2015.
\newblock Contextualized sarcasm detection on {T}witter.
\newblock In \emph{Proc.\ of the 9th International Conference on Web and Social
  Media}, pages 574--577, Oxford, UK.

\bibitem[{Benton et~al.(2016)Benton, Arora, and Dredze}]{Benton2016}
A.~Benton, R.~Arora, and M.~Dredze. 2016.
\newblock Learning multiview embeddings of {T}witter users.
\newblock In \emph{Proc.\ of {ACL}}, pages 14--19, Berlin, Germany.

\bibitem[{Chen et~al.(2016)Chen, Sun, Tu, Lin, and Liu}]{Chen2016}
H.~Chen, M.~Sun, C.~Tu, Y.~Lin, and Z.~Liu. 2016.
\newblock Neural sentiment classification with user and product attention.
\newblock In \emph{Proc.\ of {EMNLP}}, pages 1650--1659, Austin, TX.

\bibitem[{Cheng et~al.(2015)Cheng, Danescu-Niculescu-Mizil, and
  Leskovec}]{Cheng2015}
J.~Cheng, C.~Danescu-Niculescu-Mizil, and J.~Leskovec. 2015.
\newblock Antisocial behavior in online discussion communities.
\newblock In \emph{Proc.\ of the International {AAAI} Conference on Web and
  Social Media}, pages 61--70, Oxford University, England.

\bibitem[{Cho et~al.(2014)Cho, van Merrienboer, Gulcehre, Bahdanau, Bougares,
  Schwenk, and Bengio}]{Cho2014}
K.~Cho, B.~van Merrienboer, C.~Gulcehre, D.~Bahdanau, F.~Bougares, H.~Schwenk,
  and Y.~Bengio. 2014.
\newblock Learning phrase representations using {RNN} encoder--decoder for
  statistical machine translation.
\newblock In \emph{EMNLP}, pages 1724--1734, Doha, Qatar.

\bibitem[{Dadvar et~al.(2013)Dadvar, Trieschnigg, Ordelman, and
  de~Jong}]{Dadvar2013}
M.~Dadvar, D.~Trieschnigg, R.~Ordelman, and F.~de~Jong. 2013.
\newblock Improving cyberbullying detection with user context.
\newblock In \emph{ECIR}, pages 693--696, Moscow, Russia.

\bibitem[{Glorot and Bengio(2010)}]{Glorot2010}
X.~Glorot and Y.~Bengio. 2010.
\newblock Understanding the difficulty of training deep feedforward neural
  networks.
\newblock In \emph{Proc.\ of the International Conference on Artificial
  Intelligence and Statistics}, pages 249--256, Sardinia, Italy.

\bibitem[{Kingma and Ba(2015)}]{Kingma2015}
D.~P. Kingma and J.~Ba. 2015.
\newblock Adam: {A} method for stochastic optimization.
\newblock In \emph{ICLR}, San Diego, CA.

\bibitem[{Lee et~al.(2014)Lee, Yang, and Rim}]{Lee2014}
J.-T. Lee, M.-C. Yang, and H.-C. Rim. 2014.
\newblock Discovering high-quality threaded discussions in online forums.
\newblock \emph{Journal of Computer Science and Technology}, 29(3):519--531.

\bibitem[{Li et~al.(2016)Li, Galley, C.~Brockett, Spithourakis, Gao, and
  Dolan}]{Li2016}
J.~Li, M.~Galley, Chris C.~Brockett, G.~Spithourakis, J.~Gao, and B.~Dolan.
  2016.
\newblock A persona-based neural conversation model.
\newblock In \emph{Proc.\ of {ACL}}, pages 994--1003, Berlin, Germany.

\bibitem[{van~der Maaten and Hinton(2008)}]{Maaten2008}
L.~J.~P. van~der Maaten and G.~E. Hinton. 2008.
\newblock Visualizing data using t-{SNE}.
\newblock \emph{Journal of Machine Learning Research}, 9:2579--2605.

\bibitem[{Mikolov et~al.(2013)Mikolov, Yih, and Zweig}]{Mikolov2013c}
T.~Mikolov, W.-t. Yih, and G.~Zweig. 2013.
\newblock Linguistic regularities in continuous space word representations.
\newblock In \emph{NAACL-HLT}, pages 746--751, Atlanta, GA.

\bibitem[{Napoles et~al.(2017)Napoles, Pappu, and Tetreault}]{Napoles2017b}
C.~Napoles, A.~Pappu, and J.~Tetreault. 2017.
\newblock Automatically identifying good conversations online (yes, they do
  exist!).
\newblock In \emph{Proc.\ of the International {AAAI} Conference on Web and
  Social Media}.

\bibitem[{Pavlopoulos et~al.(2017{\natexlab{a}})Pavlopoulos, Malakasiotis, and
  I.}]{Pavlopoulos2017}
J.~Pavlopoulos, P.~Malakasiotis, and Androutsopoulos I. 2017{\natexlab{a}}.
\newblock Deep learning for user comment moderation.
\newblock In \emph{Proc.\ of the {ACL} Workshop on Abusive Language Online},
  Vancouver, Canada.

\bibitem[{Pavlopoulos et~al.(2017{\natexlab{b}})Pavlopoulos, Malakasiotis, and
  I.}]{Pavlopoulos2017emnlp}
J.~Pavlopoulos, P.~Malakasiotis, and Androutsopoulos I. 2017{\natexlab{b}}.
\newblock Deeper attention to abusive user content moderation.
\newblock In \emph{EMNLP}, Copenhagen, Denmark.

\bibitem[{Pennington et~al.(2014)Pennington, Socher, and
  Manning}]{Pennington2014}
J.~Pennington, R.~Socher, and C.~Manning. 2014.
\newblock Glo{V}e: Global vectors for word representation.
\newblock In \emph{EMNLP}, pages 1532--1543, Doha, Qatar.

\bibitem[{Waseem and Hovy(2016)}]{Waseem2016}
Z.~Waseem and D.~Hovy. 2016.
\newblock Hateful symbols or hateful people? {P}redictive features for hate
  speech detection on {T}witter.
\newblock In \emph{Proc.\ of {NAACL} Student Research Workshop}, pages 88--93,
  San Diego, CA.

\bibitem[{Wulczyn et~al.(2017)Wulczyn, Thain, and Dixon}]{Wulczyn2017}
E.~Wulczyn, N.~Thain, and L.~Dixon. 2017.
\newblock Ex machina: Personal attacks seen at scale.
\newblock In \emph{WWW}, pages 1391--1399, Perth, Australia.

\bibitem[{Zhang et~al.(2016)Zhang, Gong, Wu, Huang, and Huang}]{Zhang2016}
Q.~Zhang, Y.~Gong, J.~Wu, H.~Huang, and X.~Huang. 2016.
\newblock Retweet prediction with attention-based deep neural network.
\newblock In \emph{Proc.\ of the International on Conference on Information and
  Knowledge Management}, pages 75--84, Indianapolis, IN.

\end{thebibliography}
\bibliographystyle{emnlp_natbib}
\end{document}